%% file: main.tex
\documentclass[conference]{svproc}
\usepackage{times}

\input{header.tex}
\usepackage{makecell}
\usepackage[numbers]{natbib}
\usepackage{multicol}
\usepackage[bookmarks=true]{hyperref}
\usepackage[font=small,labelfont=bf]{caption}
\title{Time Integrating Articulated Body Dynamics Using Position-Based Collocation Methods\vspace{-15px}}
\titlerunning{Time Integrating Articulated Body Dynamics Using Position-Based Collocation Method}
\author{\vspace{-10px}Zherong Pan\inst{1} \and Dinesh Manocha\inst{1}}
\authorrunning{Zherong Pan, Dinesh Manocha et al.}
\institute{University of North Carolina, North Carolina NC 27514, USA,\\
\email{\{zherong,dm\}@cs.unc.edu}\vspace{-20px}}

\begin{document}
\maketitle
\thispagestyle{empty}
\pagestyle{empty}

\begin{abstract}
We present a new time integrator for articulated body dynamics. We formulate the governing equations of the dynamics using only the position variables and then recast the position-based articulated dynamics as an optimization problem. Our reformulation allows us to integrate the dynamics in a fully implicit manner without computing high-order derivatives. Therefore, under arbitrarily large timestep sizes, we observe highly stable behaviors using an off-the-shelf numerical optimizer. Moreover, we show that the accuracy of our time integrator can increase by using a  high-order collocation method. We show that each iteration of optimization has a complexity of $\ATMOST(N)$ using the Quasi-Newton method or $\ATMOST(N^2)$ using Newton's method, where $N$ is the number of links of the articulated model. Finally, our method is highly parallelizable and can be accelerated using a Graphics Processing Unit (GPU). We highlight the efficiency and stability of our method on different benchmarks and compare the performance with prior articulated body dynamics simulation methods based on the Newton-Euler equation. Using a larger timestep size, our method achieves up to $4$ times speedup on a single-core CPU. With GPU acceleration, we observe an additional $3-6$ times speedup over a 4-core CPU.
\end{abstract}

\input{introduction.tex}
\input{related.tex}
\input{velocity.tex}
\input{position.tex}
\input{algorithm.tex}
\input{results.tex}
\input{conclusion.tex}

\appendix
\ifarxiv
\input{appenA.tex}
\input{appenB.tex}
\fi
\bibliographystyle{splncs03}
\bibliography{template}
\end{document}

%% file: header.tex
\usepackage{amsmath,amsfonts,amssymb}
\usepackage{prettyref}
\usepackage{mathrsfs}
\usepackage{graphicx}
\usepackage{wrapfig}
\usepackage{xcolor}
\usepackage{subfig}
\usepackage{multirow}
\usepackage{booktabs}
\usepackage{algorithm}
\usepackage{algpseudocode}
\algnewcommand{\LineComment}[1]{\State \(\triangleright\) #1}
\algdef{SE}[DOWHILE]{Do}{doWhile}{\algorithmicdo}[1]{\algorithmicwhile\ #1}
\newcommand*{\colorboxed}{}
\def\colorboxed#1#{%
  \colorboxedAux{#1}%
}
\newcommand*{\colorboxedAux}[3]{%
  \begingroup
    \colorlet{cb@saved}{.}%
    \color#1{#2}%
    \boxed{%
      \color{cb@saved}%
      #3%
    }%
  \endgroup
}

\newrefformat{Fig}{Figure~\ref{#1}}
\newrefformat{fig}{Figure~\ref{#1}}
\newrefformat{par}{Section~\ref{#1}}
\newrefformat{appen}{Appendix~\ref{#1}}
\newrefformat{sec}{Section~\ref{#1}}
\newrefformat{sub}{Section~\ref{#1}}
\newrefformat{table}{Table~\ref{#1}}
\newrefformat{alg}{Algorithm~\ref{#1}}
\newrefformat{Alg}{Algorithm~\ref{#1}}
\newrefformat{Def}{Definition~\ref{#1}}
\newrefformat{Thm}{Theorem~\ref{#1}}
\newrefformat{step}{Step~\ref{#1}}
\newrefformat{ln}{Line~\ref{#1}}
\newrefformat{eq}{Equation~\ref{#1}}
\newrefformat{pb}{Problem~\ref{#1}}
\newrefformat{it}{Item~\ref{#1}}
\newrefformat{te}{Term~\ref{#1}}
\def\Eqref Eq:#1:{\eqref{eq:#1}}
\newrefformat{Eq}{Equation~\Eqref#1:}

\newcommand{\E}[1]{\mathbf{#1}}
\newcommand{\TE}[1]{\textbf{#1}}

\newcommand{\FPP}[2]{\frac{\partial{#1}}{\partial{#2}}}
\newcommand{\FPPROW}[2]{{\partial{#1}}/{\partial{#2}}}
\newcommand{\FPPT}[2]{\frac{\partial^2{#1}}{\partial{#2}^2}}
\newcommand{\FPPTROW}[2]{{\partial^2{#1}}/{\partial{#2}^2}}

\newcommand{\FPPTTTROW}[2]{{\partial^3{#1}}/{\partial{#2}^3}}
\newcommand{\FPPTT}[3]{\frac{\partial^2{#1}}{\partial{#2}\partial{#3}}}
\newcommand{\FPPTTROW}[3]{{\partial^2{#1}}/{\partial{#2}\partial{#3}}}

\newcommand{\TWO}[2]{\left(\begin{array}{cc}{#1} & {#2}\end{array}\right)}
\newcommand{\TWOC}[2]{\left(\begin{array}{c}#1 \\ #2\end{array}\right)}
\newcommand{\TWOR}[2]{\left(\begin{array}{cc}{#1}^T & {#2}^T\end{array}\right)^T}
\newcommand{\THREE}[3]{\left(\begin{array}{ccc}{#1} & {#2} & {#3}\end{array}\right)}

\newcommand{\FOUR}[4]{\left(\begin{array}{cccc}{#1} & {#2} & {#3} & {#4}\end{array}\right)}

\newcommand{\MTT}[4]{\left(\begin{array}{cc}#1 & #2 \\ #3 & #4\end{array}\right)}

\newcommand{\argminP}{\E{argmin}\;}

\newcommand{\BODY}{\mathcal{B}}
\newcommand{\qq}{\E{q}}
\newcommand{\PP}{\E{P}}
\newcommand{\DT}{\Delta t}
\newcommand{\POT}{\E{Q}}
\newcommand{\pp}{\E{p}}
\newcommand{\RR}{\E{R}}
\newcommand{\ttt}{\E{t}}
\newcommand{\TT}{\E{T}}
\newcommand{\AAA}{\E{A}}
\newcommand{\BBB}{\E{B}}
\newcommand{\CCC}{\E{C}}
\newcommand{\DDD}{\E{D}}
\newcommand{\PPP}{\E{Proj}_\parallel}
\newcommand{\ddd}{\E{d}}
\newcommand{\GGG}{\E{E}}
\newcommand{\GGGT}{\E{F}}
\newcommand{\GGGTT}{\E{G}}
\newcommand{\ff}{\E{f}}
\newcommand{\JJ}{\E{J}}
\newcommand{\MM}{\E{M}}
\newcommand{\EE}{\E{E}}
\newcommand{\HH}{\E{H}}
\newcommand{\II}{\E{I}}
\newcommand{\CROSS}[1]{[#1]}
\newcommand{\ww}{\boldsymbol{\omega}}
\newcommand{\ATMOST}{\mathcal{O}}
\newcommand{\TR}[1]{\E{tr}\left[#1\right]}

\newif\ifarxiv
\arxivtrue
\newcommand{\ARXIVREF}[1]{\ifarxiv\prettyref{#1}\else our extended report \cite{1709.04145}\fi}

%% file: introduction.tex
\section{Introduction}\label{sec:intro}
Numerical modeling of articulated bodies is a fundamental problem in robotics. It is important in the design and evaluation of mechanisms, robot arms, and humanoid robots. Furthermore, articulated body simulators are increasingly used to evaluate a controller during reinforcement learning~\cite{Duan:2016:BDR:3045390.3045531,4058714}, to predict the future state of a robot during online control \cite{6907751}, and to satisfy the dynamics constraints for motion planners~\cite{4399305}. In all these applications, the underlying algorithms are implemented on top of dynamic simulators and may invoke these simulators thousands of times for different parameters and settings~\cite{6907751}. As a result, the performance of these applications is easily affected by these simulators' performances.

Many widely-used articulated body simulation packages~\cite{ode:2008,6907751} are based on implicit time-stepping schemes~\cite{stewart2000implicit}. These methods model the articulated body's governing equation as an ordinary differential equation (ODE) and then integrate the ODE using high-order numerical schemes. These methods can be arbitrarily accurate but require small timestep sizes. One simple strategy to improve the runtime performance is to use a large timestep size~\cite{schroeder2011coupled}. This strategy has proven successful in some applications, such as controlling humanoid robots~\cite{6386025}, where the timestep size used in a controller can be larger than that used in the underlying simulator. A key issue in using a large timestep size is ensuring that the time integrator is still stable. For example, the stable region of a semi-implicit Euler integrator shrinks as the timestep size increases~\cite{Butcher:1251813}. To time integrate an articulated body under a large timestep size, a simple and widely-used method is to use an unconditionally stable fully implicit Euler integrator \cite{Butcher:1251813}. However, in a conventional articulated body's governing dynamics equation, the use of a fully implicit Euler integrator involves a costly $\ATMOST(N^3)$ computation of high-order derivatives, where $N$ is the number of links in an articulated body.

\TE{Main Results:} We present {position-based articulated dynamics} (PBAD), a novel optimization-based algorithm for articulated body dynamics simulation. Unlike prior method \cite{stewart2000implicit}, which represents the velocity as a time derivative and evaluates this derivative analytically, our PBAD formulation represents this velocity using finite differences in the Euclidean space. This Euclidean space discretization allows us to represent all the physical variables as functions of positions. As a result, we can integrate the system implicitly without high-order derivatives. In addition, we show that numerical simulation under our PBAD framework can be recast as a numerical optimization. Therefore, our time integrator is stable under an arbitrarily large timestep size because a numerical optimizer can ensure that the energy value decreases during each iteration through line-search \cite{Liu1989} or trust region limitation \cite{levenberg1944method}. Solving these unconstrained minimization problems requires evaluating the energy gradient and/or Hessian and solving a linear system of size $\ATMOST(N)$. To this end, we use techniques similar to well-known forward- and inverse-dynamics algorithms~\cite{Featherstone:2007:RBD:1324846} and show that the necessary energy gradient and Hessian information can be computed within $\ATMOST(N)$ and $\ATMOST(N^2)$. Finally, we show that the accuracy of PBAD time integrator can be improved by approximating the velocity using high-order polynomials, leading to a high-order collocation method \cite{Guo2009}. 

We have implemented our algorithm and evaluated the performance on many articulated models with $10-200$ DOFs. Compared with a conventional semi-implicit Euler integrator, our PBAD simulator achieves up to $4$ times overall speedup with a serial implementation running on a single-core CPU. Finally, all the operations in our unconstrained energy minimization are inherently parallel and we accelerate the simulation on a GPU to obtain $3-6$ times additional speedup over a 4-core CPU, as shown in \prettyref{sec:GPU}. 

The rest of the paper is organized as follows. We first review conventional Lagrangian articulated body dynamics in \prettyref{sec:vel} and then introduce our PBAD formulation in \prettyref{sec:pos}. Next, in \prettyref{sec:alg}, we present some algorithmic and numeric analysis of our method. Finally, we compare our method with an earlier method \cite{stewart2000implicit} on a set of classic benchmarks used by \cite{ode:2008,6907751} in \prettyref{sec:res}. We also show some applications in online/offline control algorithms in \prettyref{sec:res}.

%% file: related.tex
\section{Related Work}\label{sec:related}
We give a brief overview of previous work in articulated body dynamics, time-integration schemes, and position-based dynamics.

\subsection{Articulated Body Dynamics}
Articulated body dynamic simulation is a classic, well-studied problem in robotics. Some methods~\cite{1608023,Deul2014,CGF:CGF12322} focus on articulated bodies with general constraints, where the configurations of articulated bodies are represented using maximal coordinates. However, tree-structured articulated bodies represented using minimal coordinates have received the most attention. Very efficient algorithms~\cite{murray1994mathematical,Featherstone:2007:RBD:1324846} have been developed for forward/inverse-dynamics and these are key steps in a dynamics simulator. These algorithms have been further accelerated using divide-and-conquer \cite{doi:10.1177/02783649922066619}, adaptivity \cite{gayle2006adaptive}, and GPU-parallelism~\cite{7847363,8000600}.

\subsection{Time Integration Schemes}
A time integrator predicts the future configuration of an articulated body given its current configuration. Time integrators vary in their computational cost, stability, and accuracy (see~\cite{Butcher:1251813,CNM:CNM963} for a review). Widely-used integrators in articulated body simulators~\cite{ode:2008,6907751}, such as explicit high-order Runge-Kutta schemes, are linear multistep methods for ODE, which requires small timestep sizes. Compared with explicit schemes, implicit Runge-Kutta schemes have better stability, some of which are also known as collocation methods \cite{ascher1998computer}. Collocation methods approximate the locus of configuration using high-order polynomials. Unlike these general-purpose integrators, special integrators such as Lie-group integrators \cite{Kobilarov:2009:LGI:1516522.1516527} and variational integrators \cite{lee2016linear} can be developed to respect the Lie group structure of articulated bodies, resulting in desirable conservative properties in linear/angular momentum and the Hamiltonian.

\subsection{Position-Based Dynamics (PBD)}
Our method is inspired by the recent advances in PBD in computer graphics (see \cite{egt.20151045} for a survey). PBD has been shown to be stable under arbitrarily large timestep sizes and is preferred for interactive applications such as game engines. PBD algorithms have been developed for various dynamics systems such as fluid bodies, deformable bodies, and rigid bodies \cite{Deul2014}. In computer graphics, however, rigid bodies are represented using maximal coordinates while in our PBAD formulation, we use minimal coordinates. We have also extended conventional second-order PBD discretizations to arbitrarily high-order collocation methods. The connection between PBD and optimization-based integrators is revealed in \cite{Bouaziz:2014:Projective} and later refined in \cite{7164346,Narain2016}. 

%% file: velocity.tex
\section{Background: Lagrangian Articulated Body Dynamics}\label{sec:vel}
We briefly review the conventional articulated body dynamics formulation under generalized coordinates (see \cite{murray1994mathematical} for more details). Throughout our derivation, we assume that there is only one rigid body. The more general case of multiple rigid bodies can be derived by a concatenation of equations for each rigid body. The configuration of a rigid body $\BODY$ is parameterized by generalized coordinates, $\qq$. $|\qq|$ is the number of DOFs and is proportional to the number of links, $N$. For an arbitrary point $\pp\in\BODY$ in the body-fixed frame of reference, its corresponding position in a global frame of reference is:
\begin{eqnarray*}
\PP(\qq)=\RR(\qq)\pp+\ttt(\qq),
\end{eqnarray*}
where $\RR$ is a global rotation and $\ttt$ is a global translation. The dynamics of $\BODY$ is governed by the following equation:
\begin{eqnarray}
\label{eq:DYN}
\int_{\pp\in\BODY} \FPP{\PP(\qq)}{\qq}^T\left[\rho\ddot{\PP}(\qq)-\ff\right]d\pp=0,
\end{eqnarray}
where $\ff$ are the internal/external forces on $\pp$ and $\rho$ is the mass density. If we analytically evaluate the second derivative in \prettyref{eq:DYN}, we arrive at the following well-known equation:
\begin{eqnarray}
\label{eq:DYN_EXP}
\JJ^T\MM\JJ\ddot{\qq}+\left[\JJ^T\MM\dot{\JJ}+\JJ^T\MTT{0}{}{}{\CROSS{\ww}}\MM\JJ\right]\dot{\qq}-\JJ^T\ff=0,
\end{eqnarray}
where we have $\dot\RR=\CROSS{\ww}\RR$, $\JJ=\TWOR{\FPPROW{\ww}{\qq}}{\FPPROW{\ttt}{\qq}}$, $\MM$ being the $6\times6$ mass matrix. From \prettyref{eq:DYN_EXP}, we can formulate a discrete version to predict the next configuration $\TWO{\qq_{k+1}}{\dot{\qq}_{k+1}}$ from the current configuration $\TWO{\qq_k}{\dot{\qq}_k}$. Here we use subscript to denote timestep index, i.e. $\qq_k$ is $\qq$ at time instance $k\DT$. To this end, several widely-used articulated body simulators \cite{6907751,ode:2008} use a semi-implicit Euler scheme:
\begin{eqnarray}
\label{eq:DYN_UPDATE}
\frac{\dot{\qq}_{k+1}-\dot{\qq}_k}{\DT}=\left[\JJ_k^T\MM_k\JJ_k\right]^{-1}\left[\JJ_k^T\ff_k-\left(\JJ_k^T\MM_k\dot{\JJ}_k+\JJ_k^T\MTT{0}{}{}{\CROSS{\ww_k}}\MM_k\JJ_k\right)\dot{\qq}_k\right].
\end{eqnarray}
The above scheme usually works well for a small timestep size (usually smaller than $0.01$s), but its stability under large timestep size is not guaranteed. This is due to the explicit velocity update in \prettyref{eq:DYN_UPDATE}, i.e. the right-hand side of \prettyref{eq:DYN_UPDATE} is at timestep $k$. One common method for achieving better stability under a large timestep size is to use the fully implicit Euler scheme by replacing $\TWO{\qq_k}{\dot{\qq}_k}$ in the right-hand side of \prettyref{eq:DYN_UPDATE} with $\TWO{\qq_{k+1}}{\dot{\qq}_{k+1}}$ and solving for $\qq_{k+1}$ using an iterative algorithm. A widely-used iterative algorithm is the (Quasi)-Newton method, which has been used to stably simulate deformable and fluid bodies \cite{schroeder2011coupled}. However, there are two difficulties in using the (Quasi)-Newton method for fully implicit integration:
\begin{itemize}
\item The (Quasi)-Newton method requires the derivatives of the right-hand side of \prettyref{eq:DYN_UPDATE} with respect to $q_{k+1}$, which involves third-order derivatives, $\FPPTTTROW{\RR}{\qq}$ and $\FPPTTTROW{\ttt}{\qq}$, the evaluation complexity of which is $\ATMOST(N^3)$.
\item The implicit integrator solves a system of nonlinear equations for which even (Quasi)-Newton method could fail to converge under large timestep sizes \cite{7164346}.
\end{itemize}

%% file: position.tex
\section{Position-based Articulated Body Dynamics}\label{sec:pos}
In this section, we present our PBAD formulation, which overcomes some of the problems found in prior time integrators. We notice that, from \prettyref{eq:DYN}, the acceleration of $\PP$ is evaluated analytically to derive \prettyref{eq:DYN_EXP}, which involves up to second-order derivatives. However, if we use a finite difference approximation of $\ddot{\PP}$ directly from \prettyref{eq:DYN}, the analytic derivatives can be eliminated, allowing us to perform a (Quasi)-Newton method without evaluating $\FPPTTTROW{\RR}{\qq}$ and $\FPPTTTROW{\ttt}{\qq}$. For example, if we use second-order finite difference approximation, \prettyref{eq:DYN} becomes:
\begin{eqnarray}
\label{eq:PBAD}
\int_{\pp\in\BODY} \FPP{\PP(\qq_{k+1})}{\qq_{k+1}}^T\left[\rho\frac{\PP(\qq_{k+1})-2\PP(\qq_{k})+\PP(\qq_{k-1})}{\DT^2}-\ff(\PP(\qq_{k+1}))\right]d\pp=0.
\end{eqnarray}
Corresponding to \prettyref{eq:DYN} under the conventional formulation, \prettyref{eq:PBAD} is the governing equation under our PBAD formulation. Note that \prettyref{eq:PBAD} converges to \prettyref{eq:DYN} as $\DT\to0$. \prettyref{eq:PBAD} takes a similar form to the governing equations in previous PBD methods \cite{Narain2016,Hahn:2012:RP:2185520.2185568} for simulating deformable bodies but is expressed for articulated bodies under minimal coordinates. We can now argue that \prettyref{eq:PBAD} overcomes the two difficulties. First, if we use the Newton's method to solve \prettyref{eq:PBAD}, we only need to evaluate derivatives up to the second-order, i.e. $\FPPTROW{\RR}{\qq}$ and $\FPPTROW{\ttt}{\qq}$. Moreover, we will show in \prettyref{sec:alg} that, if we use the Quasi-Newton method, only first-order derivatives are needed without modifying the final solutions. Second, the convergence difficulty of the (Quasi)-Newton method under a very large timestep size can be fixed by reformulating \prettyref{eq:PBAD} as an energy minimization problem:
\begin{eqnarray}
\label{eq:EQ}
\EE(\qq)\triangleq\int_{\pp\in\BODY}
\left[\frac{\rho}{2\DT^2}\|\PP(\qq_{k+1})-2\PP(\qq_{k})+\PP(\qq_{k-1})\|^2+\POT(\PP(\qq_{k+1}))\right]d\pp,
\end{eqnarray}
where $\POT$ is the potential energy for a position-dependent conservative force $\ff$. Such a reformulation allows us to use an off-the-shelf, gradient-based optimizer to solve for $\qq_{k+1}=\argminP\EE(\qq)$. These optimizers use line-search \cite{Liu1989} or trust region limitations \cite{levenberg1944method} to ensure that each iteration gets the solution closer to a local minima of $\EE(\qq)$, i.e. the correct $\qq_{k+1}$. Although $\EE(\qq)$ in \prettyref{eq:EQ} still involves an integral over $\BODY$, we can derive its analytic form.

\subsection{High-Order Position-Based Collocation Method}
One advantage of using \prettyref{eq:DYN} is that one could use a general linear multistep method (see~\cite{Butcher:1251813}) to achieve a variable-order of accuracy. We show that our PBAD formulation can also have such flexibility by modifying a high-order collocation method \cite{ascher1998computer}. A collocation method approximates the locus of the configuration of $\BODY$ using high-order polynomials. Note that, in \prettyref{eq:PBAD}, we assume that, for any $\pp\in\BODY$, its trajectory in the period of time $[(k-1)\DT,(k+1)\DT]$ is determined by the three collocation points $\PP(\qq_{k-1}),\PP(\qq_k),\PP(\qq_{k+1})$ and a collocation method assumes that $\pp$ follows a polynomial curve passing through all the collocation points. For example, in \prettyref{eq:EQ}, we can fit a quadratic curve from the three points so that it is a second-order collocation method. 

To develop higher-order methods, we introduce additional collocation points in between timesteps ($\PP(\qq_{k+\alpha_1}),\cdots,\PP(\qq_{k+\alpha_{N-2}})$) for an $K$th-order method, where $0<\alpha_1<\cdots<\alpha_{K-1}=1$. We fit an $K$th-order polynomial for any $\pp\in\BODY$ from the $K+1$ collocation points $\PP_*\triangleq\THREE{\PP(\qq_{k-1+\alpha_{K-2}})}{\cdots}{\PP(\qq_{k+\alpha_{K-1}})}$. The $K$th-order polynomial takes the following form:
\begin{eqnarray*}
\PP(t)\triangleq\PP_*\HH\FOUR{1}{t}{\cdots}{t^K}^T\quad
\ddot\PP(t)\triangleq\PP_*\HH''\FOUR{1}{t}{\cdots}{t^K}^T,
\end{eqnarray*}
where $\HH,\HH''$ are the polynomial basis matrices. We call this a position-based collocation method. A key difference between a position-based collocation method and a conventional collocation method \cite{ascher1998computer} is that we fit polynomials for $\PP$ instead of $\qq$. In other words, we assume that any $\pp\in\BODY$ follows a polynomial curve in the Cartesian workspace instead of the configuration space. By plugging $\PP(t)$ into \prettyref{eq:DYN}, we obtain:
\begin{eqnarray}
\label{eq:HOC}
\int_{\pp\in\BODY} \FPP{\PP(\qq_i)}{\qq_i}^T\left[\rho\ddot{\PP}(i\DT)-\ff(\PP(\qq_i))\right]d\pp=0\quad\forall i=k+\alpha_1,\cdots,k+\alpha_{K-1}.
\end{eqnarray}
from which we can solve for $\qq_{*}=\THREE{\qq_{k+\alpha_1}}{\cdots}{\qq_{k+\alpha_{K-1}}}$ simultaneously. Given a set of collocation points, we have completed our high-order formulation of PBAD. In practice, we follow \cite{Guo2009} and use the roots of the Legendre polynomials as our collocation points. In other words, suppose $L_{K-2}(x)$ is the $(K-2)$th-order Legendre polynomial of the first kind, then $L_{K-2}(2\alpha_i-1)=0$ for $i=1,\cdots,K-2$. Note that, although \prettyref{eq:HOC} allows fully implicit integration without high-order derivatives, it does not have a corresponding energy form like \prettyref{eq:EQ}. However, we can still govern the convergence of a gradient-based optimizer using the following energy form:
\begin{eqnarray}
\label{eq:EQHOC}
\EE(\qq_*)=\sum_{i=k+\alpha_1}^{i=k+\alpha_{K-1}}\|\int_{\pp\in\BODY} \FPP{\PP(\qq_i)}{\qq_i}^T\left[\rho\ddot{\PP}(i\DT)-\ff(\PP(\qq_i))\right]d\pp\|^2,
\end{eqnarray}
where we solve for all the $\qq_{*}$ from $\qq_{*}=\argminP\EE(\qq_*)$. The high-order position-based collocation method (\prettyref{eq:EQHOC}) is more general than its second order counterpart (\prettyref{eq:EQ}) because $\ff$ is not integrated to get $\POT$, allowing $\ff$ to be non-conservative. Further, \prettyref{eq:EQHOC} still allows simulation in a fully implicit manner without computing third-order derivatives.

%% file: algorithm.tex
\section{Optimization Algorithm}\label{sec:alg}
In this section, we introduce the algorithm that performs numerical simulations under our PBAD formulation. During the timestep $k$, an implementation of our PBAD articulated body simulator calls a gradient-based optimizer to solve $\qq_{*}=\argminP \EE(\qq_*)$, where $\EE$ takes the form of \prettyref{eq:EQ} for second-order collocation methods and conservative force models and $\EE$ takes the form of \prettyref{eq:EQHOC} for high-order collocation methods or non-conservative force models. Each timestep is an iterative algorithm whose complexity is not a constant. However, we can analyze the complexity of each iteration and profile the number of iterations empirically.
\input{EEval.tex}

Our objective functions involve both inertial and potential energy terms. Since the concrete form of potential energy $\POT$ is application-dependent, we focus on the inertial term. Values and derivatives of most widely-used potential energies, such as the gravitational energy, can be evaluated in $\ATMOST(N)$ or $\ATMOST(N^2)$ and the complexity of algorithm is dominated by the inertial term. During each iteration, we evaluate the value and the partial derivatives of $\EE$, which involve an integral over $\BODY$. We can evaluate this integral analytically. Note that $\EE$ in \prettyref{eq:EQ} is a linear combination of the following term:
\begin{eqnarray}
\label{eq:ATOM}
\II(\qq_a,\qq_b)=\int_{\pp\in\BODY}\PP(\qq_a)^T\PP(\qq_b)d\pp,
\end{eqnarray}
with different $(a,b)$-pairs, as shown in \ARXIVREF{appen:VDeriv}. Similarly, $\EE$ in \prettyref{eq:EQHOC} is a linear combination of \prettyref{eq:ATOM}'s partial derivatives. \prettyref{eq:ATOM} can be evaluated analytically as:
\begin{small}
\begin{eqnarray*}
\II(\qq_a,\qq_b)&=&\left[\TT(\qq_a)^T\TT(\qq_b)\right]:\left[\int_{\pp\in\BODY}\TWOC{\pp}{1}\TWOC{\pp}{1}^Td\pp\right]-1,
\end{eqnarray*}
\end{small}
where the integrals on the right-hand side can be precomputed. We have used contract symbols such that $\AAA:\BBB=\TR{\AAA^T\BBB}$ and we have used homogeneous coordinates:
\begin{small}
\begin{eqnarray*}
\TT(\qq)=\MTT{\RR(\qq)}{\ttt(\qq)}{}{1}.
\end{eqnarray*}
\end{small}

\begin{table}[ht]
\setlength{\tabcolsep}{5pt}
\vspace{-20px}
\begin{center}
\begin{tabular}{|c|c|c|}
\toprule
\diaghead{\theadfont OptimizerObjective}{Optimizer}{Objective} & \prettyref{eq:EQ} & \prettyref{eq:EQHOC}  \\
\midrule
LM     & 
$\II(\qq_a,\qq_b),\FPP{\II(\qq_a,\qq_b)}{\qq_b},\FPPTT{\II(\qq_a,\qq_b)}{\qq_a}{\qq_b}$ &
$\FPP{\II(\qq_a,\qq_b)}{\qq_b},\FPPT{\II(\qq_a,\qq_b)}{\qq_b},\FPPTT{\II(\qq_a,\qq_b)}{\qq_a}{\qq_b}$\\
\midrule
LBFGS  & 
$\II(\qq_a,\qq_b),\FPP{\II(\qq_a,\qq_b)}{\qq_b}$ &
$\FPP{\II(\qq_a,\qq_b)}{\qq_b},\FPPT{\II(\qq_a,\qq_b)}{\qq_b},\FPPTT{\II(\qq_a,\qq_b)}{\qq_a}{\qq_b}$\\
\bottomrule
\end{tabular}
\end{center}
\vspace{-5px}
\caption{\label{table:deriv} The variables required by different optimizers using different objective functions. Since high-order methods are more frequently used, we use \prettyref{eq:EQHOC} as our objective function in most cases.}
\vspace{-25px}
\end{table} 
To solve $\qq_{*}$, we consider two optimizers, LBFGS \cite{Liu1989} and LM \cite{levenberg1944method}. Given an objective function $\EE(\qq_*)$, each iteration of LBFGS computes a gradient, $\FPPROW{\EE(\qq_*)}{\qq_*}$, and updates $\qq_*$ using a line-search along the gradient direction to ensure the decrease of $\EE(\qq_*)$. The cost of an LBFGS iteration is dominated by the computation of the gradient which takes $\ATMOST(N^2)$ in the case of \prettyref{eq:EQHOC} and $\ATMOST(N)$ in the case of \prettyref{eq:EQ}. Unlike LBFGS, each iteration of LM computes a gradient, $\FPPROW{\EE(\qq_*)}{\qq_*}$, and a $\JJ^T\JJ$-approximate Hessian, $\JJ^T\JJ(\EE(\qq_*))$, and updates $\qq_*$ using the Newton's method:
\begin{eqnarray*}
\qq_*\gets\qq_*-\left[\JJ^T\JJ(\EE(\qq_*))+\lambda\E{I}\right]^{-1}\FPP{\EE(\qq_*)}{\qq_*},
\end{eqnarray*}
where $\lambda$ is tuned to ensure the decrease of $\EE(\qq_*)$. To compute the $\JJ^T\JJ$-approximate Hessian, our objective function must be a sum-of-squares, as is the case with \prettyref{eq:EQHOC}, or an integral-of-squares, as is the case with \prettyref{eq:EQ}. The cost of an LM iteration is dominated by solving a linear system of size $|\qq|\times|\qq|$, and is $\ATMOST(N^3)$ assuming a general linear solver. 

The two optimization algorithms require different partial derivatives of $\II(\qq_a,\qq_b)$ (up to second order) during each iteration, as illustrated in \prettyref{table:deriv}. The values and derivatives of $\II(\qq_a,\qq_b)$ can be computed efficiently using the adjoint method, which results in algorithms similar to the forward/inverse dynamic algorithms in \cite{Featherstone:2007:RBD:1324846}. To introduce these algorithms, we need notations for multiple rigid bodies. We assume that we have $N$ rigid bodies $\BODY^1,\cdots,\BODY^N$, where the parent of $\BODY^i$ is $\BODY^{i-1}$. We use superscripts to denote body indices. For each $\BODY^i$, we denote its transformation as $\TT^i$ and we have $\TT^i=\TT^{i-1}\tt_{i-1}^i$. With these notations, $\II(\qq_a,\qq_b)=\sum_i\II^i(\qq_a,\qq_b)$ becomes the summation of all the bodies. We compute $\II(\qq_a,\qq_b)$ and $\FPPROW{\II(\qq_a,\qq_b)}{\qq_b}$ within $\ATMOST(N)$ using \prettyref{alg:DERIV}. We compute $\FPPTROW{\II(\qq_a,\qq_b)}{\qq_b}$ within $\ATMOST(N^2)$ using \prettyref{alg:HESS1} and we compute $\FPPTTROW{\II(\qq_a,\qq_b)}{\qq_a}{\qq_b}$ within $\ATMOST(N^2)$ using \prettyref{alg:HESS2}.

\subsection{Algorithm Complexity of High-Order Collocation Methods}
Compared with second-order collocation method that only optimizes $\qq_{k+1}$, high-order collocation methods optimize multiple $\qq$ in $\qq_*$. In addition, we can only use \prettyref{eq:EQHOC} as the objective function. The cost of each iteration of the optimization algorithm is dominated by computing the matrix $\FPPTTROW{\II(\qq_a,\qq_b)}{\qq_a}{\qq_b}$. This matrix has size $|\qq_*|\times|\qq_*|$ and can be decomposed into $(K-2)\times(K-2)$ blocks of size $|\qq|\times|\qq|$. Each block is computed using \prettyref{alg:HESS2} and takes $\ATMOST(N^2)$, so that the computation of the entire $|\qq_*|\times|\qq_*|$ matrix takes $\ATMOST((K-2)^2N^2)$.

\input{GPUImpl.tex}

%% file: EEval.tex
\setlength{\textfloatsep}{10pt}
\begin{algorithm}[ht]
\caption{Compute $\II(\qq_a,\qq_b),\FPP{\II(\qq_a,\qq_b)}{\qq_b}$ using adjoint method within $\ATMOST(N)$. Here $\AAA$ is a $4\times4$ matrix. Note that \prettyref{ln:AAA} is $\ATMOST(1)$ because $\FPPROW{\TT_{i-1}^i(\qq_b)}{\qq_b}$ is non-zero only at entries corresponding to the $i$th link.}\label{alg:DERIV}
\begin{algorithmic}[1]
\State $\TT^0(\qq_a)\gets\E{I},\TT_0^1(\qq_a)\gets\E{I}$
\State $\TT^0(\qq_b)\gets\E{I},\TT_0^1(\qq_b)\gets\E{I}$
\State $\II(\qq_a,\qq_b)\gets 0$
\For {$i=1,\cdots,N$}\Comment{$\ATMOST(N)$ forward pass}
\State $\TT^i(\qq_a)\gets \TT^{i-1}(\qq_a)\TT_{i-1}^i(\qq_a)$ 
\State $\TT^i(\qq_b)\gets \TT^{i-1}(\qq_b)\TT_{i-1}^i(\qq_b)$
\State $\II(\qq_a,\qq_b)\gets \II(\qq_a,\qq_b)+\II^i(\qq_a,\qq_b)$
\EndFor
\State $\AAA\gets 0$,$\FPP{\II(\qq_a,\qq_b)}{\qq_b}\gets 0$\Comment{$\AAA$ is $4\times4$ matrix}
\For {$i=N,\cdots,1$}\Comment{$\ATMOST(N)$ backward pass}
\State $\AAA\gets \AAA+\FPP{\II^i(\qq_a,\qq_b)}{\TT^i(\qq_b)}$
\State $\FPP{\II(\qq_a,\qq_b)}{\qq_b}\gets \FPP{\II(\qq_a,\qq_b)}{\qq_b}+(\TT^{i-1}\FPP{\TT_{i-1}^i(\qq_b)}{\qq_b}):\AAA$\Comment{$\ATMOST(1)$}
\label{ln:AAA}
\State $\AAA\gets \AAA\TT_{i-1}^i(\qq_b)^T$
\EndFor
\end{algorithmic}
\end{algorithm}

\setlength{\textfloatsep}{10pt}
\begin{algorithm}[ht]
\caption{Compute $\FPPT{\II(\qq_a,\qq_b)}{\qq_b}$ using adjoint method within $\ATMOST(N^2)$. Here $\AAA,\BBB$ are $4\times4$ matrices.}\label{alg:HESS1}
\begin{algorithmic}[1]
\LineComment{Same forward pass as \prettyref{alg:DERIV}.}

\State $\AAA\gets 0,\FPPT{\II(\qq_a,\qq_b)}{\qq_b}\gets 0$
\For {$i=N,\cdots,1$}\Comment{$\ATMOST(N^2)$ backward pass}
\State $\AAA\gets \AAA+\FPP{\II^i(\qq_a,\qq_b)}{\TT^i(\qq_a)}$
\State $\FPPT{\II(\qq_a,\qq_b)}{\qq_b}\gets \FPPT{\II(\qq_a,\qq_b)}{\qq_b}+(\TT^{i-1}\FPPT{\TT_{i-1}^i(\qq_b)}{\qq_b}):\AAA$\Comment{$\ATMOST(1)$}
\State $\BBB\gets \AAA\FPP{\TT_{i-1}^i(\qq_b)}{\qq_b}^T$
\For {$j=i-1,\cdots,1$}
\State $\FPPT{\II(\qq_a,\qq_b)}{\qq_b}\gets\FPPT{\II(\qq_a,\qq_b)}{\qq_b}+
(\TT^{j-1}\FPP{\TT_{j-1}^j(\qq_b)}{\qq_b}):\BBB$\Comment{$\ATMOST(1)$}
\State $\BBB\gets \BBB\TT_{j-1}^j(\qq_b)^T$
\EndFor
\State $\AAA\gets \AAA\TT_{i-1}^i(\qq_b)^T$
\EndFor
\end{algorithmic}
\end{algorithm}

\setlength{\textfloatsep}{10pt}
\begin{algorithm}[ht]
\caption{Compute $\FPPTT{\II(\qq_a,\qq_b)}{\qq_a}{\qq_b}$ using adjoint method within $\ATMOST(N^2)$. Here $\GGG,\GGGT,\GGGTT$ are $4\times4\times4\times4$ tensors, $\AAA,\BBB,\CCC,\DDD$ are $4\times4$ matrices, and we use double contraction such that $\AAA:\GGG:\BBB=\sum_{xyzw}\left[\GGG_{xyzw}\BBB_{wz}\right]$ and we have $\AAA:\CCC\GGG\DDD:\BBB=\AAA\CCC:\GGG:\DDD\BBB$. Finally, we define $\GGG_{xyzw}=\FPPTTROW{\II(\qq_a,\qq_b)}{\TT_{xy}(\qq_a)}{\TT_{zw}(\qq_b)}$.}\label{alg:HESS2}
\begin{algorithmic}[1]
\LineComment{Same forward pass as \prettyref{alg:DERIV}.}

\State $\GGG\gets 0,\FPPTT{\II(\qq_a,\qq_b)}{\qq_a}{\qq_b}\gets 0$
\For {$i=N,\cdots,1$}\Comment{$\ATMOST(N^2)$ backward pass}
\State $\GGG\gets \GGG+\FPPTT{\II^i(\qq_a,\qq_b)}{\TT^i(\qq_a)}{\TT^i(\qq_b)},
\GGGT\gets \GGG, \GGGTT\gets \GGG$
\For {$j=i,\cdots,1$}
\State $\FPPTT{\II(\qq_a,\qq_b)}{\qq_a}{\qq_b}\gets\FPPTT{\II(\qq_a,\qq_b)}{\qq_a}{\qq_b}+$
\State $(\TT^{i-1}(\qq_a)\FPP{\TT_{i-1}^i(\qq_a)}{\qq_a}):\GGGT:(\TT^{j-1}(\qq_b)\FPP{\TT_{j-1}^j(\qq_b)}{\qq_b})^T+$
\State $(\TT^{j-1}(\qq_a)\FPP{\TT_{j-1}^j(\qq_a)}{\qq_a}):\GGGTT:(\TT^{i-1}(\qq_b)\FPP{\TT_{i-1}^i(\qq_b)}{\qq_b})^T$\Comment{$\ATMOST(1)$}
\State $\GGGT\gets \GGGT\TT_{j-1}^j(\qq_b)^T, \GGGTT\gets \TT_{j-1}^j(\qq_a)\GGGTT$
\EndFor
\State $\GGG\gets \TT_{i-1}^i(\qq_a)\GGG\TT_{i-1}^i(\qq_b)^T$
\EndFor
\end{algorithmic}
\end{algorithm}

%% file: GPUImpl.tex
\subsection{GPU Parallelization}\label{sec:GPU}
\begin{figure}[ht]
\begin{center}
\includegraphics[width=0.7\textwidth]{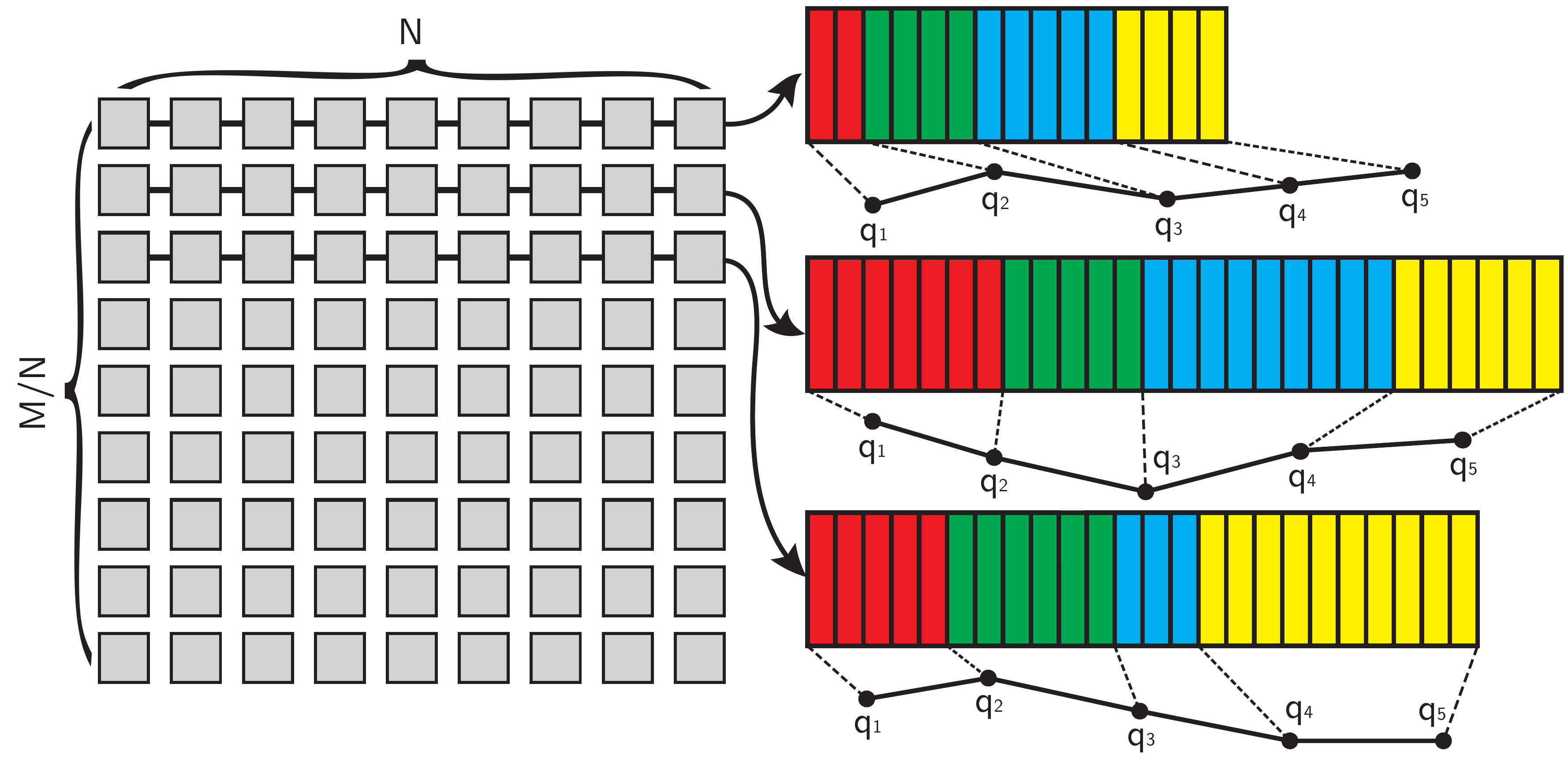}
\vspace{-15px}
\end{center}
\caption{\label{fig:GPU} An illustration of our GPU implementation. The GPU has $M$ cores, each illustrated as a gray box on the left. We use a workgroup of $N$ cores (black arrow) to simulate one trajectory. In this illustration, we compute 3 trajectories that each have $4$ timesteps ($\qq_2,\cdots,\qq_5$). During each call to the GPU, instead of finishing the entire LM optimization, we compute just one iteration of the LM optimization (colored block on the right) so that all the workgroups are running the same computation and no starvation will happen. Different timesteps are illustrated using blocks of different colors. For example, it takes $2$ iterations to compute $\qq_2$ in the first trajectory and $7$ iterations to compute $\qq_2$ in the second trajectory (red block).}
\vspace{-10px}
\end{figure}

Our PBAD formulation is designed to be GPU-friendly. Simulating rigid bodies on a GPU has been previously studied~\cite{8000600,7847363}. These methods formulate forward/inverse dynamics algorithms as GPU-scan operations. Our GPU implementation deviates from \cite{8000600,7847363} in two ways. First, our implementation is intended to be used for modeling predictive control~\cite{6386025} and reinforcement learning~\cite{Duan:2016:BDR:3045390.3045531}, where we need to generate multiple trajectories at once. This fact provides more opportunities for parallelism. Second, our algorithm is iterative and the number of iterations performed during each timestep tends to be different. In practice, an implementation that runs each timestep in a separate thread could result in starvation, where threads finishing early are waiting for other threads. As a result, we parallelize each iteration of an optimization instead of each timestep. This mechanism is illustrated in \prettyref{fig:GPU}.

We choose the LM algorithm in our GPU implementation. Each iteration of LM involves computing $\FPP{\II(\qq_a,\qq_b)}{\qq_b},\FPPTROW{\II(\qq_a,\qq_b)}{\qq_b},\FPPTTROW{\II(\qq_a,\qq_b)}{\qq_a}{\qq_b}$ according to \prettyref{table:deriv} and then using a linear system solver. The serial computation of $\FPPTROW{\II(\qq_a,\qq_b)}{\qq_b}$ and $\FPPTTROW{\II(\qq_a,\qq_b)}{\qq_a}{\qq_b}$ takes $\ATMOST(N^2)$, which can be costly. We introduce an additional fine-grain parallelism by using a GPU workgroup of $N$ cores to reduce the complexity of computing the partial derivatives to $\ATMOST(N)$ using algorithms in \ARXIVREF{appen:GPU}. With the same workgroup of $N$ cores, the complexity of the GPU linear solver is reduced to $\ATMOST(N^2)$ using parallel Cholesky factorization \cite{galoppo2005lu}. As a result, a GPU with $M$ cores can simulate $\lfloor M/N\rfloor$ trajectories in parallel and the complexity of each iteration is dominated by the linear solver, i.e. is $\ATMOST(N^2)$. This method is suitable for modern commodity GPUs with the number of cores $M\gg N$. 

Finally, in \prettyref{sec:res}, we will show that widely used external force models such as frictional contact forces and fluid drag forces can be formulated as integrable energies, $\POT$, whose values and derivatives can be computed in a similar manner to the inertial terms computed in this section. Putting them together, our method can be used to model the complex locomotion tasks in \cite{Duan:2016:BDR:3045390.3045531}, such as swimming, walking, and jumping.

%% file: results.tex
\section{Results \& Applications}\label{sec:res}
In this section, we evaluate the performance of our formulation on several benchmarks. 

\subsection{Comparison}
Throughout this section, we compare our formulation with conventional formulations based on \prettyref{eq:DYN_EXP} and integrated using the Runge-Kutta method \cite{Butcher:1251813}. The same algorithm is implemented in \cite{ode:2008,6907751}. Note that the definition of order of integration is different for the Runge-Kutta method and the position-based collocation method. The position-based collocation method of order $K$ has accuracy similar to that of the Runge-Kutta method of order $K-1$. All experiments are performed on a single desktop machine with a 4-core CPU (Intel i7-4790 3.6G) and a 3584-core GPU (Nvidia Titan-X), i.e. $M=3584$.

\begin{figure}[ht]
\vspace{-15px}
\begin{center}
\includegraphics[width=0.99\textwidth]{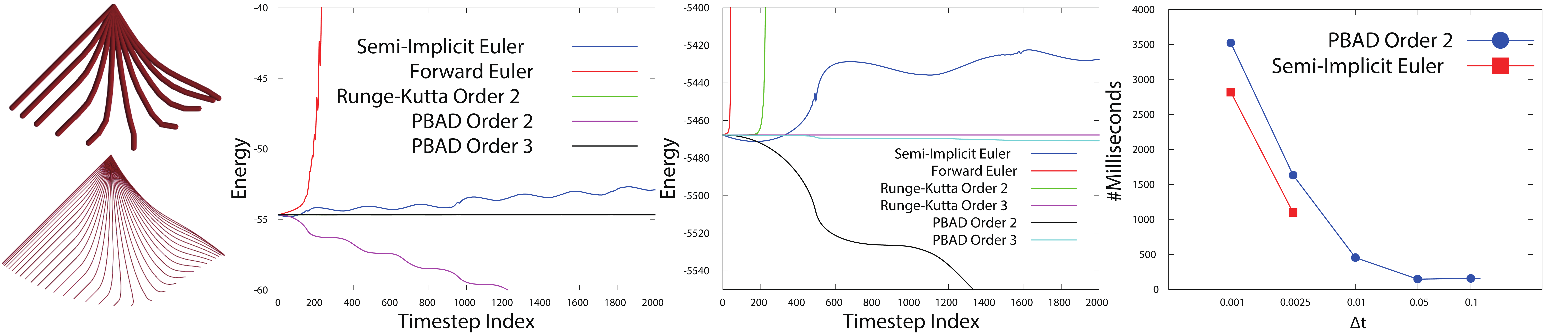}
\put(-300,60){(a)}
\put(-300, 5){(b)}
\put(-220,-2){(a)}
\put(-120,-2){(b)}
\put( -20,-2){(c)}
\end{center}
\vspace{-15px}
\caption{\label{fig:SWING} (a): We plot the total kinetic+potential energy over time during a standard simulation of a 10-link chain (20-DOF) that swings downward. Forward Euler integrator for the Newton-Euler equation and semi-implicit Euler integrator are not stable. Being fully implicit, our second-order PBAD solver is stable but quickly loses energy. By increasing the order by one, both the second-order Runge-Kutta and our third-order PBAD solver preserve energy very well. (b): For the more challenging task of a 100-link chain (200-DOF) that swings downward, even the second-order Runge-Kutta method is not stable and we have to use the third-order Runge-Kutta method for better energy preservation. Our second-order PBAD solver is stable but quickly loses energy. Our third-order PBAD solver preserves energy very well. (c): We compare the total computational time for generating a $10$s trajectory of a 10-link chain swinging down using a second-order collocation method for PBAD and a semi-implicit Euler integrator for a conventional formulation. PBAD is $1.5-2.1$ times slower at a small timestep size and up to $4$ times faster at a large timestep size, such as $0.05$s.}
\vspace{-5px}
\end{figure}
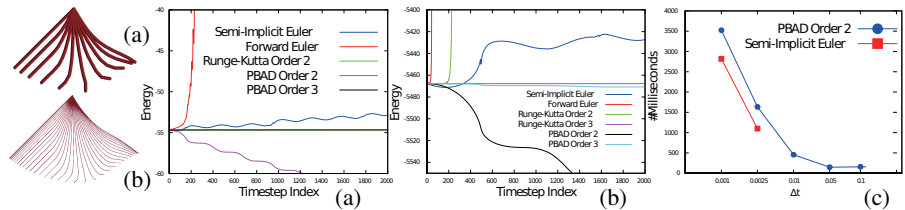
\TE{Energy Preservation:} We compare the accuracy of time integrators for our PBAD formulation and conventional formulation. In \prettyref{fig:SWING} (a), we plot the total kinetic+potential energy over time during a standard simulation of a 10-link chain (20-DOF) that swings downward (the same benchmark was used in~\cite{gayle2006adaptive}). The timestep size is $0.0025$s. We can see that PBAD is very stable and continuously loses energy (\prettyref{fig:SWING} (a) purple). In contrast, low-order explicit integrators such as forward Euler and semi-implicit Euler are not stable. For better accuracy, we can increase the order of integration by one, resulting in a much better performance in terms of energy preservation. In \prettyref{fig:SWING} (b), we redo the experiment for a 100-link chain (200-DOF). This is more challenging and low-order explicit integrators are more unstable. The Runge-Kutta method for the Newton-Euler equation is stable at the third order. Although our second-order PBAD solver suffers a fast energy loss, increasing the order by one can significantly improve accuracy.

\TE{Timestep Size:} In \prettyref{fig:SWING} (c), we compare the total computational time for generating a $10$s trajectory of a 10-link chain that swings downward using a second-order collocation method for PBAD and a semi-implicit Euler integrator for a conventional formulation. Each timestep of PBAD integration is costlier because multiple iterations of computations are needed to ensure the optimizer converges. For example, when we use timestep sizes of $0.001$s and $0.0025$s, the total computational time of the PBAD integrator is $1.5-2.1$ times that of the semi-implicit integrator. However, the PBAD integrator can be more efficient under a larger timestep size, while $0.0025$s is the largest timestep size that works for the semi-implicit Euler integrator. At a timestep size of $0.05$s, the total computational time of the PBAD integrator is $0.21$ times that of the semi-implicit integrator, leading to a $4$ times speedup.

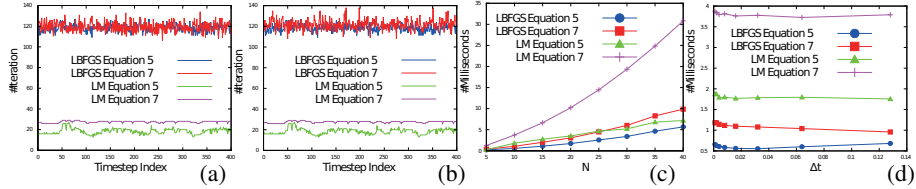
\begin{figure}[ht]
\vspace{-15px}
\begin{center}
\includegraphics[width=0.99\textwidth]{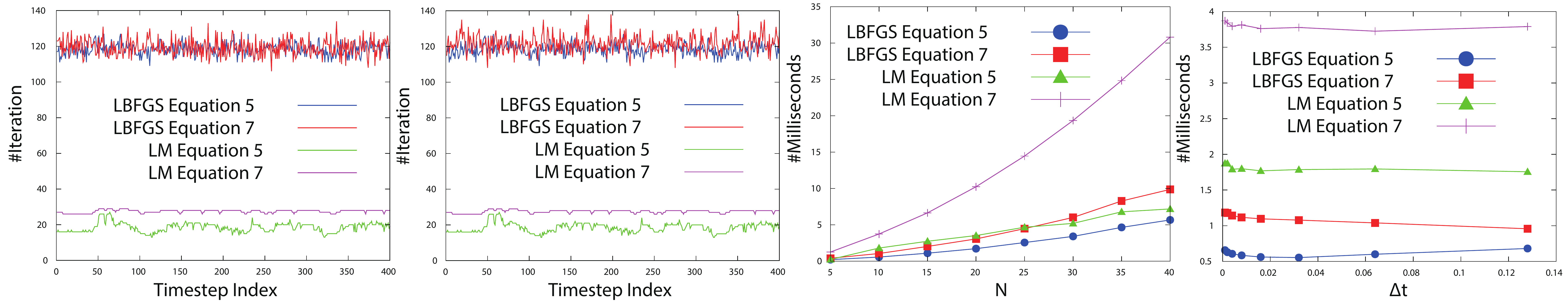}
\put(-270,-2){(a)}
\put(-180,-2){(b)}
\put(-100,-2){(c)}
\put(-10 ,-2){(d)}
\end{center}
\vspace{-15px}
\caption{\label{fig:OPTProf} We compare the performance of the two optimization algorithms (LM and LBFGS) during the simulation of a 10-link (20-DOF) (a) and a 40-link (80-DOF) chain (b) with a large timestep size of $0.05$s. The number of iterations used by LBFGS is much larger than that used by LM, although each iteration of LBFGS is cheaper. In addition, the number of iterations is almost independent of the number of links, $N$. (c): We plot the average time to finish one step of the simulation against the number of links, $N$. LBFGS is comparable to LM in terms of computational time and the computational time grows almost linearly with $N$ in the range of $N=10-40$. (d): We plot the average time to finish one step of the simulation against the timestep size, $\Delta t$. PBAD can be used with very large timestep sizes and we tested $\Delta t=0.001,0.002,0.004,0.008,0.016,0.032,0.064,0.128$s. The computation time for each timestep is almost invariant to $\Delta t$.}
\vspace{-5px}
\end{figure}

\TE{Optimization Algorithm:} We compare the performance of the two optimization algorithms (LM and LBFGS) on CPU. \prettyref{fig:OPTProf} (a, b) shows that, LBFGS generally takes $10$ times more iterations than LM. In addition, PBAD integration performed using \prettyref{eq:EQHOC} as the objective function will require more iterations to converge than when using \prettyref{eq:EQ}. Moreover, the numbers of iterations used by both algorithms are independent of the number of links, $N$. Considering the number of iterations as an invariant, the cost of LM grows as $\ATMOST(N^3)$ and the cost of LBFGS grows as $\ATMOST(N)$ on CPU. However, \prettyref{fig:OPTProf} (c) shows that, when the number of links $N<40$, the total computational time grows almost linearly. In particular, using LM to optimize \prettyref{eq:EQHOC} is costlier than other choices. \prettyref{fig:OPTProf} (c) also shows that the computation times of LBFGS and LM are comparable. Finally, PBAD can be used with very large timestep sizes, such as $\Delta t=0.128$s, shown in \prettyref{fig:OPTProf} (d), and the average time to compute each timestep is almost invariant to the timestep size. Therefore, large timestep sizes lead to a reduction in total computation time but they also lead to a higher rate of numerical dissipation.

\begin{figure}[t]
\begin{center}
\includegraphics[width=0.99\textwidth]{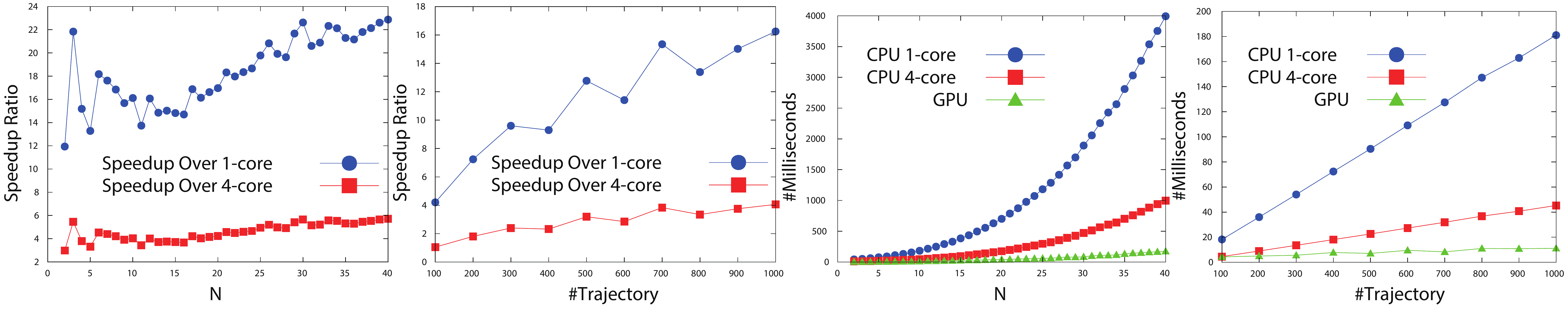}
\put(-270,-2){(a)}
\put(-180,-2){(b)}
\put(-100,-2){(c)}
\put(-10 ,-2){(d)}
\end{center}
\vspace{-15px}
\caption{\label{fig:CPUGPU} We compare the performance of CPU and GPU in simulating a chain swinging benchmark. (a): We plot the speedup against the number of links, $N$. The speedup increases with $N$ and the maximal speedup over a 4-core CPU is $6$ times. (b): When $N=10$, we plot the speedup against the number of trajectories. The speedup also increases with the number of trajectories and the maximal speedup is $4$ times. (c): We plot the total computational time against the number of links, $N$, for generating $100$ trajectories of $10$ timesteps each. When $N=40$, the $100$ trajectories can be generated in less than $1$s on GPU. (d): We plot the total computational time against the number of trajectories.}
\vspace{-5px}
\end{figure}
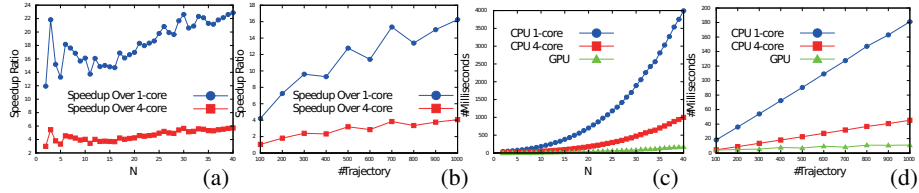
\TE{GPU Acceleration:} We compare the performance of our PBAD formulation on CPU and GPU. Our GPU implementation only provides acceleration when multiple trajectories are simulated simultaneously for different initial conditions, which is the case with many online/offline control algorithms such as \cite{Duan:2016:BDR:3045390.3045531,6386025}. In \prettyref{fig:CPUGPU} (a, b), we show the speedup of our GPU implementation over a 4-core CPU. The speedup increases with both the number of links and the number of trajectories to be computed. The speedup is between 3-6 times. The total computational time for generating $100$ trajectories of $10$ timesteps each is plotted in \prettyref{fig:CPUGPU} (c). On GPU, generating these trajectories takes less than $1$s for $N\leq40$. Finally, in \prettyref{fig:CPUGPU} (d), we plot the total computational time against the number of trajectories to be computed when $N=10$. Note that our GPU has $3584$ cores and we can compute $\lfloor M/N\rfloor=358$ trajectories in parallel. Therefore, when the number of trajectories increases from $100-300$, more GPU cores are used and the total computational time does not increase. Therefore, the green curve in \prettyref{fig:CPUGPU} (d) is almost flat.

\begin{figure}[t]
\begin{center}
\includegraphics[width=0.99\textwidth]{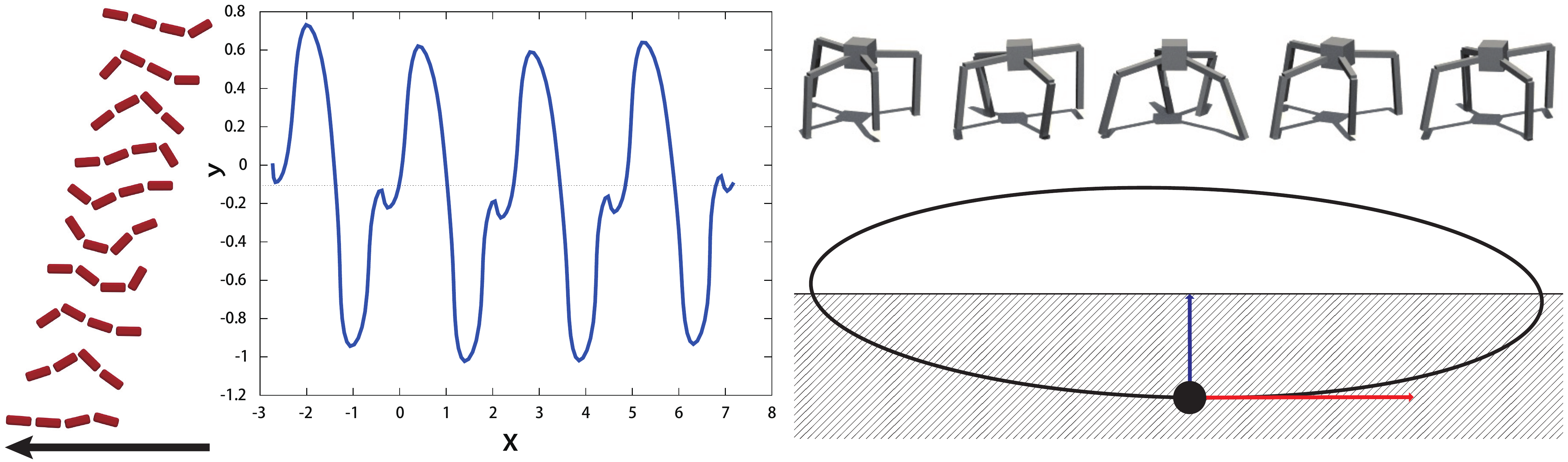}
\put(-310,10){(a)}
\put(-20,100){(b)}
\put(-20, 55){(c)}
\put(-70, 26){\textcolor{blue}{$\ddd\left(\PP(\qq_{k+1})\right)$}}
\put(-30, 10){\textcolor{red}{$\PPP$}}
\end{center}
\vspace{-15px}
\caption{\label{fig:RL} (a): A 4-linked swimmer is trained to swim forward using CMA-ES. Some optimized swimming gaits and the locus of the center-of-mass are shown. (b): A 4-linked spider is trained to walk forward. Some optimized walking gaits are shown. The notations used by our frictional contact force model, \prettyref{eq:FRI}, are shown in (c). Our model is penalty-based. Both normal and tangential forces are related to the penetration depth $\ddd\left(\PP(\qq_{k+1})\right)$ (blue). The tangential forces (red) are modelled as a velocity damping term.}
\vspace{-5px}
\end{figure}
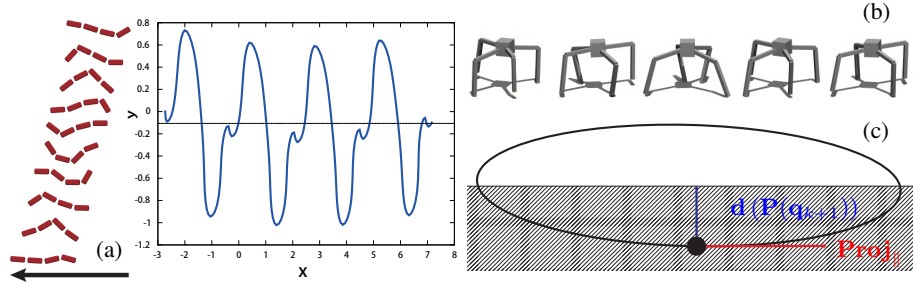
\subsection{Applications in Controller Optimization}
In our last benchmark, we use the PBAD formulation as the underlying simulator for controller optimization applications. These applications require multiple trajectories to be generated simultaneously, and our GPU implementation can provide performance improvements for them. We choose two benchmarks from \cite{Duan:2016:BDR:3045390.3045531}, swimmer and spider. 

In the swimmer benchmark, a 4-linked chain (9-DOF, including a 6-DOF rigid transformation and 3 hinge joints) is immersed underwater and is under constant fluid drag forces. To model these drag forces, we use the following formulation of potential energy in \prettyref{eq:EQ}:
\begin{small}
\begin{eqnarray*}
\POT_{drag}(\PP(\qq_{k+1}),\PP(\qq_k))=D\|\frac{\PP(\qq_{k+1})-\PP(\qq_{k})}{\DT}\|^2,
\end{eqnarray*}
\end{small}
where $D$ is the drag force coefficient. This term minimizes the velocity of $\pp$ and can be considered as a damping force model. An integral of $\POT_{drag}$ over $\BODY$ can be written as a linear combination of \prettyref{eq:ATOM} with different (a,b)-pairs as shown in \ARXIVREF{appen:VDeriv} so that its value and derivatives can be computed using the techniques discussed in \prettyref{sec:alg}. We use CMA-ES \cite{hansen1996adapting} to optimize a controller for the swimmer to move forward and the results are shown in \prettyref{fig:RL} (a), where $20$ trajectories of $1000$ timesteps with $\DT=0.05$s are sampled simultaneously on GPU during each iteration. Each iteration takes $12$s on average and the entire optimization takes $100$ iterations and about $20$min. 

In the spider benchmark, a 4-linked spider (18-DOF, including a 6-DOF rigid transformation, 4 ball joints, and 4 hinge joints) is trying to move forward on the ground. In this case, the 4-linked spider is under frictional contact forces and gravitational forces. A previous method \cite{stewart2000implicit} handles frictional contact forces using complementary conditions, which requires a sequential algorithm. To integrate the frictional contact forces with our energy minimization framework, we use a penalty-based frictional contact model by using the following potential energy in \prettyref{eq:EQ}:
\begin{small}
\begin{eqnarray}
\label{eq:FRI}
\POT_{contact}(\PP(\qq_{k+1}),\PP(\qq_k))&=&D_1\|\ddd\left(\PP(\qq_{k+1})\right)\|^2+    \\
&&D_2\|\ddd\left(\PP(\qq_{k+1})\right)\|^2\|\PPP\left(\frac{\PP(\qq_{k+1})-\PP(\qq_{k})}{\DT}\right)\|^2\nonumber.
\end{eqnarray}
\end{small}
Here $D_1$ is the normal force penalty and $\ddd$ is the penetration depth, which is positive when $\PP$ is inside obstacles and zero otherwise, as illustrated in \prettyref{fig:RL} (c). $D_2$ is the frictional force penalty and $\PPP$ is the projection matrix to the tangential directions. The integral of $\POT_{contact}$ over $\BODY$ is replaced by a summation of a set of discrete contact points. The second term on the right-hand side of \prettyref{eq:FRI} approximates frictional forces by requiring tangential velocities to be small when a point $\PP$ is inside any of the obstacles. We use policy gradient method \cite{4058714} to optimize a controller for the spider to move forward; the results are shown in \prettyref{fig:RL} (b). We use $100$ iterations of policy gradients and each iteration takes $27$s on average and the entire optimization takes less than $1$hr.

%% file: conclusion.tex
\section{Conclusion, Limitations \& Future Work}
In this paper, we present the PBAD reformulation of articulated body dynamics. Our reformulation casts the simulation as an energy minimization problem. As a result, off-the-shelf optimizers can be used to stably simulate articulated bodies under very large timestep sizes. Although each timestep of our algorithm requires more iterations than conventional methods, the overall speedup of our PBAD over conventional methods in various benchmarks is up to $4$ times under very large timestep sizes, e.g., $\Delta t=0.1$s. Furthermore, our approach is GPU friendly and can be easily parallelized. We observe an additional $3-6$ times speedup on a commodity GPU over a 4-core CPU. The parallel version of our PBAD solver can accelerate control algorithms such as model predictive control and reinforcement learning by simulating multiple trajectories simultaneously.

Our current formulation still has some limitations. First, numerical dissipation cannot totally be avoided, although we can reduce it using smaller timestep sizes or high-order collocation methods. Second, to recast the articulated body dynamics as an optimization problem and avoid high-order derivatives, we discretize the velocities in a Euclidean workspace, instead of using a Lie-Group structure \cite{lee2016linear}. As a result, our PBAD method can be less accurate compared with Lie-Group integrators. As part of future work, we would like to study various external force models that are compatible with the our PBAD formulation. A compatible force model should be stable under large timestep sizes. To this end, one method is to formulate the external force implicitly as a function of $\qq_{k+1}$ \prettyref{eq:FRI}. However, the accuracy of these force models have not been well studied.

%% file: appenA.tex
\section{Value and Derivatives of The Objective Function}\label{appen:VDeriv}
We show how to assemble objective functions from \prettyref{eq:ATOM} without considering the potential term, $\POT$. If \prettyref{eq:EQ} is used as the objective function, then we have:
\begin{eqnarray*}
\EE(\qq_{k+1})&=&\int_{\pp\in\BODY}\left[\frac{\rho}{2\DT^2}\|\PP(\qq_{k+1})-2\PP(\qq_{k})+\PP(\qq_{k-1})\|^2\right]d\pp   \\
&=&\frac{\rho}{2\DT^2}(\II(\qq_{k+1},\qq_{k+1})+4\II(\qq_{k},\qq_{k})+\II(\qq_{k-1},\qq_{k-1}) \\
&&\quad\quad\quad -4\II(\qq_{k+1},\qq_{k})-4\II(\qq_{k-1},\qq_{k})+2\II(\qq_{k+1},\qq_{k-1}))   \\
&=&\frac{\rho}{2\DT^2}(\II(\qq_{k+1},\qq_{k+1})-4\II(\qq_{k+1},\qq_{k})+2\II(\qq_{k+1},\qq_{k-1}))+const,
\end{eqnarray*}
where $const$ is independent of $\qq_{k+1}$ and does not need to be computed for forward simulation. If \prettyref{eq:EQHOC} is used as the objective function and we use a second-order collocation method, then we have:
\begin{eqnarray*}
\EE(\qq_{k+1})&=&\|\frac{\rho}{\DT^2}(\FPP{\II(\qq_{k+1},\qq_{k+1})}{\qq_{k+1}}-2\FPP{\II(\qq_{k+1},\qq_{k})}{\qq_{k+1}}+\FPP{\II(\qq_{k+1},\qq_{k-1})}{\qq_{k+1}})\|^2.
\end{eqnarray*}
Objective functions for high-order collocation methods can be derived similarly. Finally, if $\POT_{drag}$ is used as potential energy, then we have:
\begin{eqnarray*}
\int_{\pp\in\BODY}\POT_{drag}(\PP(\qq_{k+1}),\PP(\qq_k))d\pp=\frac{D}{\Delta t^2}
(\II(\qq_{k+1},\qq_{k+1})-2\II(\qq_{k+1},\qq_k)+\II(\qq_k,\qq_k)).
\end{eqnarray*}

%% file: appenB.tex
\section{Parallel Algorithms}\label{appen:GPU}
The two algorithms used by our GPU implementation. They are very similar to \prettyref{alg:DERIV}, \prettyref{alg:HESS1}, and \prettyref{alg:HESS2}. Note that the red lines in these algorithms will not lead to any writing conflict because they write to different non-zero entries.

\begin{algorithm}[h]
\caption{Compute $\II(\qq_a,\qq_b),\FPP{\II(\qq_a,\qq_b)}{\qq_b}$ in parallel using adjoint method within $\ATMOST(N)$.}\label{alg:DERIVP}
\begin{algorithmic}[1]
\State $\TT^0(\qq_a)\gets\E{I},\TT_0^1(\qq_a)\gets\E{I}$
\State $\TT^0(\qq_b)\gets\E{I},\TT_0^1(\qq_b)\gets\E{I}$
\State $\II(\qq_a,\qq_b)\gets 0,\FPP{\II(\qq_a,\qq_b)}{\qq_b}\gets 0$
\For {$i=1,\cdots,N$}\Comment{in parallel}
\State $\TT^0(\qq_a)\gets\E{I},\TT_0^1(\qq_a)\gets\E{I}$
\State $\TT^0(\qq_b)\gets\E{I},\TT_0^1(\qq_b)\gets\E{I}$
\For {$j=1,\cdots,i$}
\State $\TT^j(\qq_a)\gets \TT^{j-1}(\qq_a)\TT_{j-1}^j(\qq_a)$ 
\State $\TT^j(\qq_b)\gets \TT^{j-1}(\qq_b)\TT_{j-1}^j(\qq_b)$
\EndFor
\State $\II(\qq_a,\qq_b)\gets \II(\qq_a,\qq_b)+\II^i(\qq_a,\qq_b)$\Comment{atomic add}
\State $\AAA\gets 0$\Comment{$\AAA$ is $4\times4$ matrix}
\For {$j=N,\cdots,i+1$}
\State $\AAA\gets(\AAA+\FPP{\II^j(\qq_a,\qq_b)}{\TT^j(\qq_b)})\TT_{j-1}^j(\qq_b)^T$
\EndFor
\State $\AAA\gets\AAA+\FPP{\II^i(\qq_a,\qq_b)}{\TT^i(\qq_b)}$
\State \textcolor{red}{$\FPP{\II(\qq_a,\qq_b)}{\qq_b}\gets \FPP{\II(\qq_a,\qq_b)}{\qq_b}+(\TT^{i-1}\FPP{\TT_{i-1}^i(\qq_b)}{\qq_b}):\AAA$}\Comment{$\ATMOST(1)$}
\EndFor
\end{algorithmic}
\end{algorithm}

\begin{algorithm}[ht]
\caption{Compute $\FPPT{\II(\qq_a,\qq_b)}{\qq_b}$ in parallel using adjoint method within $\ATMOST(N)$.}\label{alg:HESS1P}
\begin{algorithmic}[1]
\LineComment{Same forward pass as \prettyref{alg:DERIVP}.}

\State $\FPPT{\II(\qq_a,\qq_b)}{\qq_b}\gets 0$
\For {$i=N,\cdots,1$}\Comment{in parallel}

\State $\AAA\gets 0$
\For {$j=N,\cdots,i+1$}
\State $\AAA\gets (\AAA+\FPP{\II^j(\qq_a,\qq_b)}{\TT^j(\qq_a)})\TT_{j-1}^j(\qq_b)^T$
\EndFor
\State $\AAA\gets \AAA+\FPP{\II^i(\qq_a,\qq_b)}{\TT^i(\qq_a)}$

\State \textcolor{red}{$\FPPT{\II(\qq_a,\qq_b)}{\qq_b}\gets \FPPT{\II(\qq_a,\qq_b)}{\qq_b}+(\TT^{i-1}\FPPT{\TT_{i-1}^i(\qq_b)}{\qq_b}):\AAA$}\Comment{$\ATMOST(1)$}
\label{ln:AA}
\State $\BBB\gets \AAA\FPP{\TT_{i-1}^i(\qq_b)}{\qq_b}^T$
\For {$j=i-1,\cdots,1$}
\State \textcolor{red}{$\FPPT{\II(\qq_a,\qq_b)}{\qq_b}\gets\FPPT{\II(\qq_a,\qq_b)}{\qq_b}+
(\TT^{j-1}\FPP{\TT_{j-1}^j(\qq_b)}{\qq_b}):\BBB$}\Comment{$\ATMOST(1)$}
\State $\BBB\gets \BBB\TT_{j-1}^j(\qq_b)^T$
\EndFor
\EndFor
\end{algorithmic}
\end{algorithm}

\begin{algorithm}[ht]
\caption{Compute $\FPPTT{\II(\qq_a,\qq_b)}{\qq_a}{\qq_b}$ in parallel using adjoint method within $\ATMOST(N)$.}\label{alg:HESS2P}
\begin{algorithmic}[1]
\LineComment{Same forward pass as \prettyref{alg:DERIVP}.}

\State $\FPPTT{\II(\qq_a,\qq_b)}{\qq_a}{\qq_b}\gets 0$
\For {$i=N,\cdots,1$}\Comment{in parallel}

\State $\GGG\gets 0$
\For {$j=N,\cdots,i+1$}
\State $\GGG\gets \TT_{j-1}^j(\qq_a)(\GGG+\FPPTT{\II^j(\qq_a,\qq_b)}{\TT^j(\qq_a)}{\TT^j(\qq_b)})\TT_{j-1}^j(\qq_b)^T$
\EndFor
\State $\GGG\gets \GGG+\FPPTT{\II^i(\qq_a,\qq_b)}{\TT^i(\qq_a)}{\TT^i(\qq_b)}, \GGGT\gets \GGG, \GGGTT\gets \GGG$

\For {$j=i,\cdots,1$}
\State \textcolor{red}{$\FPPTT{\II(\qq_a,\qq_b)}{\qq_a}{\qq_b}\gets\FPPTT{\II(\qq_a,\qq_b)}{\qq_a}{\qq_b}+$}
\State \textcolor{red}{$(\TT^{i-1}(\qq_a)\FPP{\TT_{i-1}^i(\qq_a)}{\qq_a}):\GGGT:(\TT^{j-1}(\qq_b)\FPP{\TT_{j-1}^j(\qq_b)}{\qq_b})^T+$}
\State \textcolor{red}{$(\TT^{j-1}(\qq_a)\FPP{\TT_{j-1}^j(\qq_a)}{\qq_a}):\GGGTT:(\TT^{i-1}(\qq_b)\FPP{\TT_{i-1}^i(\qq_b)}{\qq_b})^T$}\Comment{$\ATMOST(1)$}
\State $\GGGT\gets \GGGT\TT_{j-1}^j(\qq_b)^T, \GGGTT\gets \TT_{j-1}^j(\qq_a)\GGGTT$
\EndFor
\EndFor
\end{algorithmic}
\end{algorithm}